\documentclass[3p,times]{elsarticle_revised}
\usepackage{ecrc_revised}

\usepackage{xcolor}
\usepackage{float}
\usepackage{graphicx}
\usepackage{hyperref}
\usepackage{amsmath}
\usepackage{multirow}
\usepackage{graphicx}

\volume{00}

\firstpage{1}

\runauth{A. J. Varghese et al.}

\usepackage{amssymb}
\usepackage[figuresright]{rotating}

\begin{document}

\begin{frontmatter}

%% Title, authors and addresses

%% use the tnoteref command within \title for footnotes;
%% use the tnotetext command for the associated footnote;
%% use the fnref command within \author or \address for footnotes;
%% use the fntext command for the associated footnote;
%% use the corref command within \author for corresponding author footnotes;
%% use the cortext command for the associated footnote;
%% use the ead command for the email address,
%% and the form \ead[url] for the home page:
%%
% \title{Title\tnoteref{label1}}
%% \tnotetext[label1]{}
% \author{Name\corref{cor1}\fnref{label2}}
%% \ead{email address}
%% \ead[url]{home page}
%% \fntext[label2]{}
%% \cortext[cor1]{}
%% \address{Address\fnref{label3}}
%% \fntext[label3]{}

% \dochead{}
%% Use \dochead if there is an article header, e.g. \dochead{Short communication}
%% \dochead can also be used to include a conference title, if directed by the editors
%% e.g. \dochead{17th International Conference on Dynamical Processes in Excited States of Solids}

\title{TransformerG2G: Adaptive time-stepping for learning temporal graph embeddings using transformers}

%%%% author list + affiliations %%%% 
\author[1]{Alan John Varghese}
\address[1]{School of Engineering, Brown University, Providence, RI 02912, USA}

\author[2]{Aniruddha Bora}
\address[2]{Division of Applied Mathematics, Brown University, Providence, RI 02912, USA}

\author[3,4]{Mengjia Xu \corref{Corresponding author}}
\address[3]{Department of Data Science, New Jersey Institute of Technology, Newark, NJ 07102, USA}
\address[4]{Center for Brains, Minds and Machines, Massachusetts Institute of Technology, Cambridge, MA 02139, USA}

\author[1,2,5]{George Em Karniadakis}
\address[5]{Pacific Northwest National Laboratory, Richland, WA 99354, USA}

% \ead[url]{}
\fntext[Corresponding author]{Corresponding author: mx6@njit.edu}

\begin{abstract}
Dynamic graph embedding has emerged as a very effective technique for addressing diverse temporal graph analytic tasks (i.e., link prediction, node classification, recommender systems, anomaly detection, and graph generation) in various applications. Such temporal graphs exhibit heterogeneous transient dynamics, varying time intervals, and highly evolving node features throughout their evolution. Hence, incorporating long-range dependencies from the historical graph context plays a crucial role in accurately learning their temporal dynamics. In this paper, we develop a graph embedding model with uncertainty quantification, TransformerG2G,  by exploiting the advanced transformer encoder to first learn intermediate node representations from its current state ($t$) and previous context (over timestamps [$t-1, t-l$], $l$ is the length of context). Moreover, we employ two projection layers to generate lower-dimensional multivariate Gaussian distributions as each node's latent embedding at timestamp $t$. We consider diverse benchmarks with varying levels of ``novelty" as measured by the TEA (Temporal Edge Appearance) plots. Our experiments demonstrate that the proposed TransformerG2G model outperforms conventional multi-step methods and our prior work (DynG2G) in terms of both link prediction accuracy and computational efficiency, especially for high degree of novelty. Furthermore, the learned time-dependent attention weights across multiple graph snapshots reveal the development of an automatic adaptive time stepping enabled by the transformer. Importantly, by examining the attention weights, we can uncover temporal dependencies, identify influential elements, and gain insights into the complex interactions within the graph structure. For example, we identified a strong correlation between attention weights and node degree at the various stages of the graph topology evolution.

\end{abstract}


\end{frontmatter}

%% main text
\section{Introduction}

Numerous real-world datasets, such as social networks, financial networks, biological protein networks, brain networks, citation networks, disease spreading networks, and transportation networks, naturally exhibit complex ``graph-like'' structures~\cite{velivckovic2023everything, salha2021fastgae}. A graph typically consists of a set of nodes that represent entities and a set of edges that depict relationships between nodes. In the past few years, graph representation learning has gained significant attention for its crucial role in effectively analyzing complex, high-dimensional graph-structured data across various application domains, e.g., drug discovery in healthcare~\cite{yi2022graph,li2022graph}, protein structure and property prediction~\cite{jumper2021highly, fang2022geometry}, brain network analysis~\cite{xu2020new, xu2021graph}, traffic forecasting~\cite{chen2023traffic, guo2021learning, li2023dynamic}, partial differential equation learning~\cite{iakovlev2020learning, eliasof2021pde}. Moreover, graph embedding techniques have gained a lot of success due to the capability to learn highly informative latent graph representations projected in a lower-dimensional space from high-dimensional graphs, while the important and intricate graph topological properties~\cite{xu2020understanding} can be maximally preserved in the latent embedding space. The lower-dimensional graph embeddings can be readily and efficiently applied for a broad range of downstream graph analytic tasks, such as link prediction, node classification, anomaly detection, recommender systems, and graph completion. However, in practice graphs usually change over time, which can be observed by newly added or removed nodes (or edges), changing node (or edge) labels or features with heterogeneous dynamic behavior. Hence, it is more challenging to learn temporal graph embedding compared to most of the existing works on static graph embeddings~\cite{grover2016node2vec, bojchevski2018deep}. 

In this work, we first present the multi-step DynG2G \cite{xu2022dyng2g} algorithm, a simple modification to the original DynG2G that helps to incorporate temporal information while learning the embeddings. Then, we propose TransformerG2G, a deep learning model with the transformer architecture at its core, which includes the history of a node while projecting it into the embedding space. We represent embeddings as multivariate Gaussian distributions, where we learn the mean and variance associated with each node. This methodology helps to quantify the uncertainty associated with the embeddings, which is typically neglected in other existing models that consider temporal history.

In section \ref{related_works} we present a review of the related works on learning graph embeddings in dynamic graphs. In section \ref{methodology}, we define dynamic graphs. Followed by a concise overview of the original DynG2G model, and discussions on the multi-step method and the TransformerG2G model. In section \ref{experiments}, we discuss the six benchmark datasets used in this study and their implementation details. We then present the results of the multi-step DynG2G model and TransformerG2G model, and focus on the time-dependent attention matrices. In section \ref{conclusions}, we present the summary of the work, limitations, and directions for future developments.

\section{Related works} 
\label{related_works}

Numerous studies have been conducted on static graph embedding using matrix factorization~\cite{ou2016asymmetric}, random walk-based approaches~\cite{grover2016node2vec, tang2015line, perozzi2014deepwalk}, and deep neural networks~\cite{wang2016structural, bojchevski2017deep, salha2022modularity}, but a major limitation of these methods is that they cannot effectively capture the important temporal information over time. Hence, dynamic graph embedding has emerged as a crucial research area for enhancing conventional graph embedding methods. It can be divided into two main categories: continuous time graph embedding and discrete time graph embedding~\cite{xu2020understanding}. 

\textit{Continuous time graph embedding} aims to learn temporal graph embedding for a series of temporal interactions at higher time resolution~\cite{zhou2018dynamic, zuo2018embedding}. Moreover, DyRep~\cite{trivedi2019dyrep} attempted to model the continuous time graph as ``temporal point processes'' and incorporate the self-attention mechanism~\cite{vaswani2017attention} to learn temporal relationship embeddings, or using recurrent neural networks (RNNs) to model the dynamics~\cite{kumar2019predicting}. TGN~\cite{rossi2020temporal} is a more generic and computationally efficient method for temporal graph embedding, because it utilizes efficient parallel processing and a single graph attention layer to update node-wise features based on the current memory and temporal node features, which address the main problem of~\cite{xu2020inductive}. 

However, in \textit{discrete-time graph embedding}, most works attempted to learn node embedding over time as lower-dimensional point vectors, i.e., each node is represented as a vector. To capture the temporal dynamics over different time stamps, some works have been conducted based on autoencoder architecture along with transfer learning approach~\cite{goyal2018dyngem} or recurrent neural networks (RNNs)~\cite{goyal2020dyngraph2vec} to capture the dynamics over timestamps. Xu et al.~\cite{xu2022dyng2g} firstly presented the ``stochastic graph embedding'' approach (i.e., DynG2G) for temporal graphs where each graph node at a timestamp can be projected into a latent space as probabilistic density function represented by mean and variance vectors. Therefore, DynG2G provides important uncertainty quantification that facilitates effective identification of the optimal embedding size. Given the great success of graph neural networks (GNN) and graph convolutional networks (GCN) in effective graph representation learning, several recent studies introduced time-dependent graph neural network (GNN) models that combine GNNs or GCNs with RNNs or long short-term memory (LSTM) for temporal graph embedding~\cite{qu2020continuous,pareja2020evolvegcn}. Recently, there has been an increasing number of works aiming to learn low dimensional embeddings for time-evolving knowledge graphs, notably \cite{xu2020temporal,han2021learning}. In \cite{xu2020temporal}, they propose ATiSE, a model that includes temporal information into entity/relation representations by using Additive Time Series decomposition and mapping them into a space of multi-dimensional Gaussian distributions. In \cite{han2021learning}, the authors propose TANGO, a continuum model for link prediction using neural ordinary differential equations and multi-relational graph convolutional networks that encode both temporal and structural information.
Nevertheless, some of the aforementioned methods may fail to accurately capture long-term historical features, especially when the temporal behaviors of nodes or edges exhibit heterogeneous dynamics over timestamps. 

%Graph embeddings have also been used in the domain of knowledge graphs. Few sentences on the relation between knowledge graph embeddings and our project. And what are knowledge graphs? Point out the difference between their work and the work we are doing.

%Xu2020 - temporal
%-These time-unaware KGE models have limitations in reasoning over temporal KGs, as they do not consider time information, leading to the same scores for quadruples with different time stamps
%--- The proposed ATiSE model addresses these limitations by incorporating time information into entity/relation representations using Additive Time Series decomposition and mapping them into the space of multi-dimensional Gaussian distributions
%--- Experimental results demonstrate that ATiSE outperforms state-of-the-art KGE models and existing temporal KGE models on link prediction over temporal KGs.

%Han et al - neuralODE
%--- The paper proposes a novel continuum model for forecasting future links on temporal knowledge graphs using neural ordinary differential equations (ODEs) and multi-relational graph convolutional networks
%--- The authors extend the idea of neural ODEs to capture the continuous nature of dynamic multi-relational graph data and encode both temporal and structural information into continuous-time dynamic embeddings.
%--- The proposed model, called TANGO, outperforms existing approaches in future link forecasting on five benchmark datasets for temporal KG reasoning

In the past few years, the self-attention mechanism, originally developed for transformers~\cite{vaswani2017attention}, has demonstrated distinct advantages in effectively and efficiently capturing long-range dependencies with adaptive and interpretable attention weights. The mechanism has been successfully applied in diverse domains (e.g., natural language processing, computer vision, and sequence modeling). To incorporate the self-attention mechanism for dynamic graph embedding, studies in \cite{sankar2020dysat} and \cite{rossi2020temporal} adopted the self-attention mechanism to jointly encode the spatial and temporal dynamics for discrete time graphs. However, the computational efficiency of the aforementioned methods decreases quadratically when applied to large-scale graphs. More comprehensive dynamic graph embedding works may be also found in some recent surveys~\cite{goyal2020graph,ji2021survey, xu2020understanding, xie2020survey, xue2022dynamic}. 

In this paper, our main focus lies in spatio-temporal stochastic discrete-time graph embedding (improved version of our prior DynG2G~\cite{xu2022dyng2g}) that emphasizes long-range dependencies with advanced transformers. Moreover, our proposed TransformerG2G model aims to encode each graph node into a latent space as a ``probability density function''. Our approach differs significantly from the aforementioned methods, which typically encode each graph (node or edge) as a ``deterministic'' lower-dimensional point-vector. However, our TransformerG2G model takes into account the crucial ``node uncertainty information'' in the latent space, which allows for the effective characterization of a wide range of node properties within the latent space, including node neighborhood diversity and optimal embedding dimensionality.

%%%%%%%%%%%%%%%%%%%%%%%%%%%%%%%%%%%%%%%%%%%%%
\section{Methodology} 
\label{methodology}
%%%%%%%%%%%%%%%%%%%%%%%%%%%%%%%%%%%%%%%%%%%%%
\subsection{Preliminaries}

\subsubsection{Dynamic graph definition}

A dynamic graph can be generally modeled in two ways: ``discrete-time'' graph snapshots or ``continuous-time'' graph~\cite{xu2020understanding}. In our study, we specifically focused on discrete-time graphs, which involve a sequence of graph snapshots captured at different timestamps. Each snapshot represents the state of the graph at a particular point in time, whose edges and nodes may change or evolve across these discrete time points. Here, we denote a temporal graph as $\mathcal{G} = \{G_t\}_{t=1}^T$, which consists of graph snapshots across $T$ timestamps.  The graph snapshot $G_t = (V_t, E_t)$ at timestamp $t$ has a vertex set $V_t = \{ v_1, v_2, v_3, \cdots, v_{|V_t|} \} $ of $|V_t|$ nodes and an edge set $E_t = \{e_{i,j} |i,j \in |V_t| \}$. 
% 

%%%%%%%%%%%%%%%%%%%%%%%%%%%%%%%%%%%%%%%%%%%%%%%%%%%%%%%%%%%%%%%%%%%
\subsubsection{Stochastic dynamic graph embedding with uncertainty quantification (DynG2G)}

\begin{figure}[h]
    \centering
    \includegraphics[width = .94\textwidth]{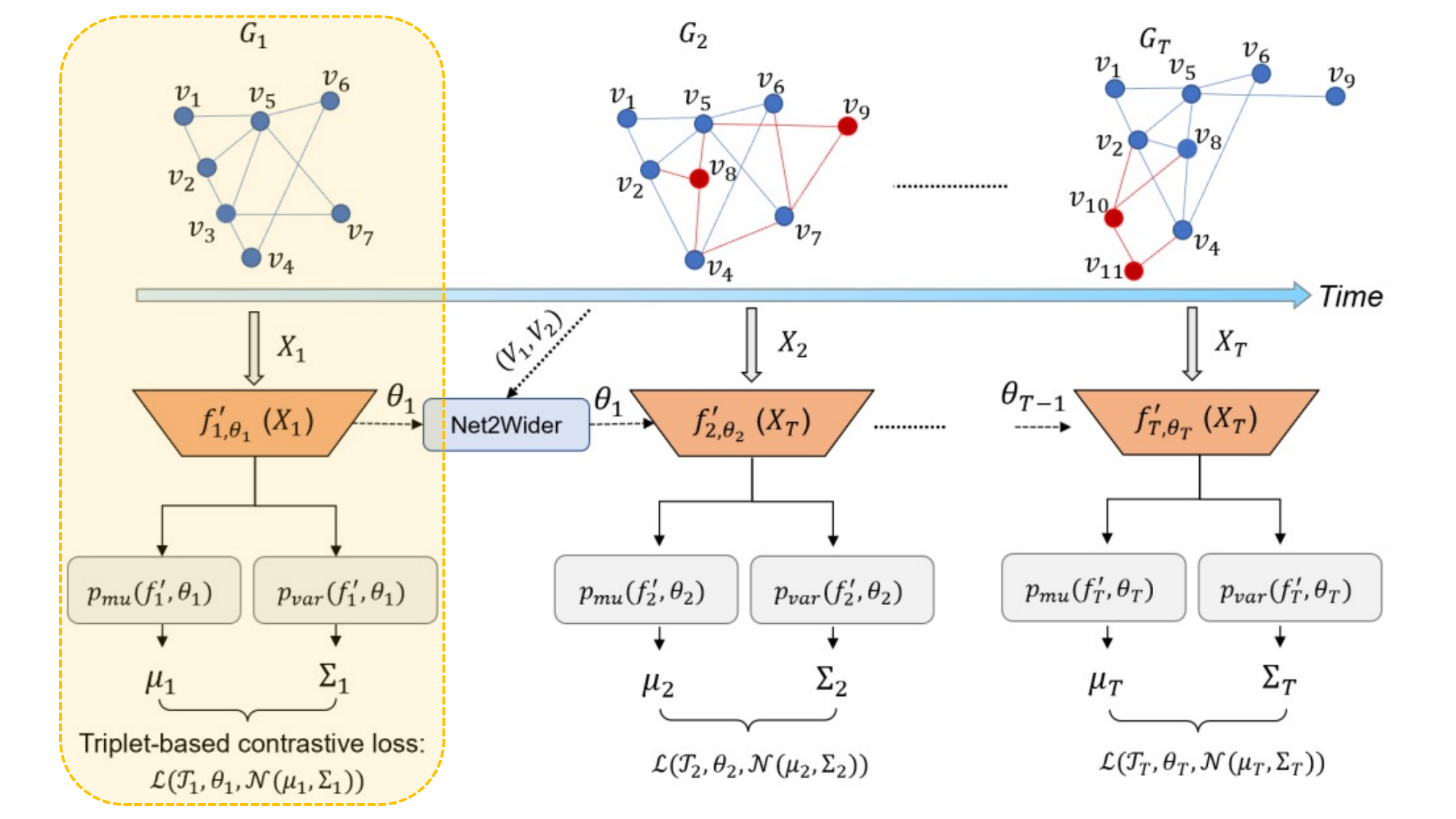}
    \caption{Main framework of our prior DynG2G model for stochastic graph embedding of temporal graphs~\cite{xu2022dyng2g}. The specific G2G model architecture for stochastic graph embedding of one static graph snapshot (yellow box) is presented in Fig.~\ref{fig:static_graph_gaussian_embedding}.}
    \label{fig:DynG2G}
\end{figure}

In our prior work~\cite{xu2022dyng2g}, we developed a temporal graph embedding method based on the DynG2G model with effective uncertainty quantification. The main framework of DynG2G is illustrated in Fig.~\ref{fig:DynG2G}. Given a dynamic graph represented by a sequence of graph snapshots ($G_1, G_2, \cdots, G_T$) at different timestamps, it first learns an intermediate representation for each node $v_i \in V_1$ in the first graph snapshot with the so-called G2G encoder~\cite{bojchevski2018deep}, and generate lower-dimensional probabilistic Gaussian embeddings in terms of the mean ($\mu_i$) and the variance ($\sigma_i$) vectors\footnote{The covariance matrix $\Sigma_i = diag(\sigma_i)$ is a square diagonal matrix with the variance vector $\sigma_i$ as its diagonal elements.}, with two additional projection heads (i.e., one single-layer linear dense network and the other one is a nonlinear single-layer dense network with ``elu'' activation function\footnote{
$\text{elu}(x) = \begin{cases}
x, & \text{if } x > 0 \\
\alpha \left(e^x - 1\right), & \text{otherwise}
\end{cases}$}). For training the next snapshot, it employs an extension of the Net2WiderNet approach~\cite{chen2015net2net} to adaptively expand the network hidden layer size based on the number of changing nodes in the next snapshot. To capture the graph dynamics, we train the encoder in the second timestamp starting with the weights transferred from the pre-trained model for the first graph snapshot embedding. Finally, the model is trained iteratively over the training dataset via optimizing a time-dependent node triplet-based contrastive loss using the Adam optimizer~\cite{kingma2014adam}. Each node is represented as a lower-dimensional multivariate Gaussian distribution in terms of the mean and variance vectors in the latent space. Hence, the uncertainty information of node embedding based on variant embedding size ($L$) in the latent space can be automatically quantified by the variance values as a function of time. A key advantage of the DynG2G~\cite{xu2022dyng2g} model is its ability to uncover the correlation between the optimal embedding size ($L_o$) and the effective dimensionality of uncertainty ($D_u$) measured by the learned embedding variances across benchmarks. The results in~\cite{xu2022dyng2g} also suggest a clear path to selecting the graph optimal embedding dimension by choosing $L_o \ge D_u$. Here, $D_u$ gives the estimation of the optimal embedding size that enables the extraction of intrinsic information from graph nodes with minimal distortion after projecting them to lower-dimensional latent space, effectively preserving graph structure properties.

% DynG2G for temporal graphs

\begin{figure}[ht]
    \centering
    \includegraphics[scale = 0.48]{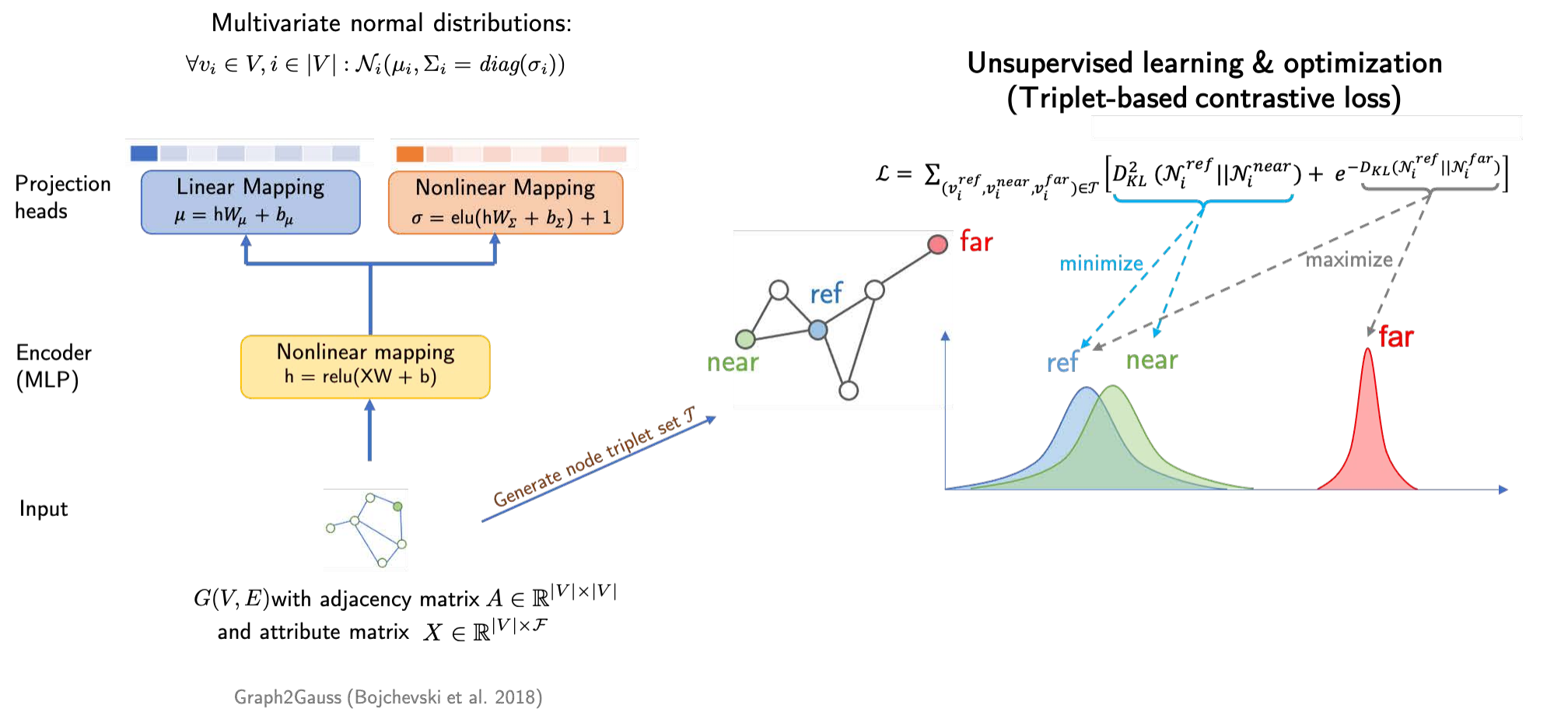}
    \caption{Gaussian node embedding for a static graph with uncertainty quantification using the Graph2Gauss method~\cite{bojchevski2018deep}. The model is trained by minimizing a triplet based contrastive loss.}
    \label{fig:static_graph_gaussian_embedding}
\end{figure}

A primary step in Fig.~\ref{fig:DynG2G} is to accurately learn graph snapshot embeddings as Gaussian distributions in the latent space. In Fig.~\ref{fig:static_graph_gaussian_embedding}, we present the specific workflow of stochastic graph embedding using the G2G model~\cite{bojchevski2018deep}. It first generates a node triplet set (\textit{ref, near, far}) using the K-hop neighborhood sampling (K = 2 or 3) technique for each node in the graph snapshot; here, \textit{ref} denotes the anchor node, \textit{near} (``small hop'' node) represents the node that is closer to the anchor node (\textit{ref}) than the \textit{far} node (``large hop'') based on the shortest distance. More details can be found in section~\ref{sec:transformers}. Essentially, (\textit{ref, near}) refers to positive node pairs, and (\textit{ref, far}) refers to negative node pairs. In order to further project the high-dimensional graph snapshot into lower-dimensional space, we input the adjacency matrix ($A$) to a single layer MLP encoder (with weights $W$ and bias $b$) followed by a ReLU activation function for achieving intermediate hidden node representations. Subsequently, we apply two projection heads to output the mean vectors ($\mu_i$) and diagonal co-variance matrices ($\Sigma_i = diag(\sigma_i)$) in the latent space, where each node is represented as a multivariate Gaussian distribution ($\mathcal{N}_i (\mu_i, \Sigma_i)$). 
The G2G model employs a node triplet-based contrastive loss ($\mathcal{L}$), structured as a squared exponential loss, as its cost function. The main objective of the G2G model optimization is to minimize the loss ($\mathcal{L}$) by decreasing the dissimilarity between the Gaussian embeddings ($\mathcal{N}_i^{\textit{ref}}, \mathcal{N}_i^{\textit{near}}$) of positive node pairs (\textit{ref, near}), and increasing the dissimilarity between the Gaussian embeddings ($\mathcal{N}_i^{\textit{ref}}, \mathcal{N}_i^{\textit{far}}$) of negative node pairs (\textit{ref, far}). The dissimilarity metric used in G2G model is the ``KL-divergence''. To this end, nodes that are proximate in the original space are encoded as joint normal distributions with a small KL-divergence, indicating a higher similarity, and vice versa. The triplet-based contrastive loss helps to maximally preserve the topological properties of the graph.
% G2G for static graphs

However, the DynG2G model has the limitation of neglecting long-term dependencies over multiple previous timestamps in the temporal graphs. In this study, we mainly focus on addressing dynamic graph embedding by incorporating long-term dependencies of graph nodes using two different methods: multi-step method and transformers~\cite{vaswani2017attention}.

\subsection{Dynamic graph embedding with multi-step methods}
\label{sec:multistep}
Several real-world datasets exhibit highly transient and possibly non-Markovian dynamics. In such datasets, including the history of the graph's temporal evolution may help to learn more meaningful graph embeddings. This is because long-term history provides crucial insights into recurring patterns, emerging trends, and potential outcomes. Our discussions in section \ref{sec:TEAplot} and \ref{sec:cosine_similarity} further elaborate on the rich temporal dynamics exhibited by the datasets considered in this work, and motivate algorithms that capture long-term dependencies. However, in DynG2G~\cite{xu2022dyng2g}, the weight initialization for a given timestamp is done solely based on the previous timestamp. To address this problem of DynG2G, we first apply a conventional multi-step method to capture temporal dynamics over timestamps. That is, we consider a linear combination of multiple previous steps to initialize the weights. To this end, we first carry out experiments with a two-step method to improve the weight initialization schema in the DynG2G model:
%  two-step method
\begin{equation}
\label{eq:twosteps}
W_{t} = \theta W_{t-1} + (1-\theta) W_{t-2}.
\end{equation}
Specifically, we initialize the weight matrix $W_t$ at timestamp $t$ using a weighted schema by $\theta$ based on the weight matrices learned from timestamps $t-1$ and $t-2$. Note that the aforementioned two-step method is equivalent to the DynG2G method when $\theta = 1$, i.e., when we use the weight matrices of just the previous timestamp to initialize the current timestamp. We also consider a three-step method, where we consider the previous three timestamps to initialize the weight matrices for the current timestamps as follows:
\begin{equation}
\label{eq:threestep}
W_t = \theta_1 W_{t-1} + \theta_2 W_{t-2} + \theta_3 W_{t-3}, \quad \text{where} \quad \theta_1 + \theta_2 + \theta_3 = 1.
\end{equation}
Here $\theta_1, \theta_2$ and  $\theta_3$ are coefficients that determine the weighting or the relative importance of the previous timestamps, and are user defined hyperparameters. A three-step method would be contain more temporal information from the past, and therefore would result in more accurate embeddings as demonstrated by the results discussed in section \ref{sec:multistep}.

The findings derived from the multi-step method offer useful insights into the benefits of increasing the temporal context for accurate and robust dynamic graph embedding learning. However, as we increase the number of previous steps considered in the multi-step method the number of coefficients  increases, resulting in a larger search space for optimal coefficients due to the curse of dimensionality. In  section~\ref{sec:transformers}, we present a major improvement on the previous DynG2G framework by replacing the G2G encoder with a transformer encoder for capturing long-term dependencies from the input sequence of graph snapshots. In contrast to the multi-step method, the self-attention mechanism in the transformer adaptively selects the weighting/relative importance of the different timestamps from the past.
% 

%%%%%%%%%%%%%%%%%%%%%%%%%%%%%%%%%%%%%%%%%%%%%%%%%%%%%%
\subsection{Dynamic graph embedding with transformers} 
\label{sec:transformers}

We consider a discrete-time temporal graph $\mathcal{G} = \{G_t\}_{t=1}^T$ with a series of graph snapshots across $T$ timestamps, where each timestamp consists of a vertex set $V_t = \{ v_1, v_2, v_3, \cdots, v_{|V_t|} \} $ of $|V_t|$ nodes and an edge set $E_t = \{e_{i,j} |i,j \in |V_t| \}$. The main goal here is to project each node from the high-dimensional dense non-Euclidean space to lower-dimensional density function space as a multivariate Gaussian distribution. This involves utilizing the temporal graphs from the current timestamp as well as previous timestamps, in order to obtain the representation for each node.

To achieve this, we propose the TransformerG2G model to learn a nonlinear mapping from a sequence corresponding to the $i$th node's history, $(v_i^{t-l}, \cdots, v_i^{t-1}, v_i^t)$, to a joint normal distribution $\mathcal{N}\left(\mu_i^t, \Sigma_i^t = \textit{diag}\left(\sigma_i^t\right)\right)$, with a transformer as the central part of its architecture to capture the temporal dynamics of the graph. The lookback $l$ is a crucial hyperparameter that denotes the number of timestamps to look back or the length of the node history considered.

\begin{figure}[ht]
    \centering
    \includegraphics[scale = .6]{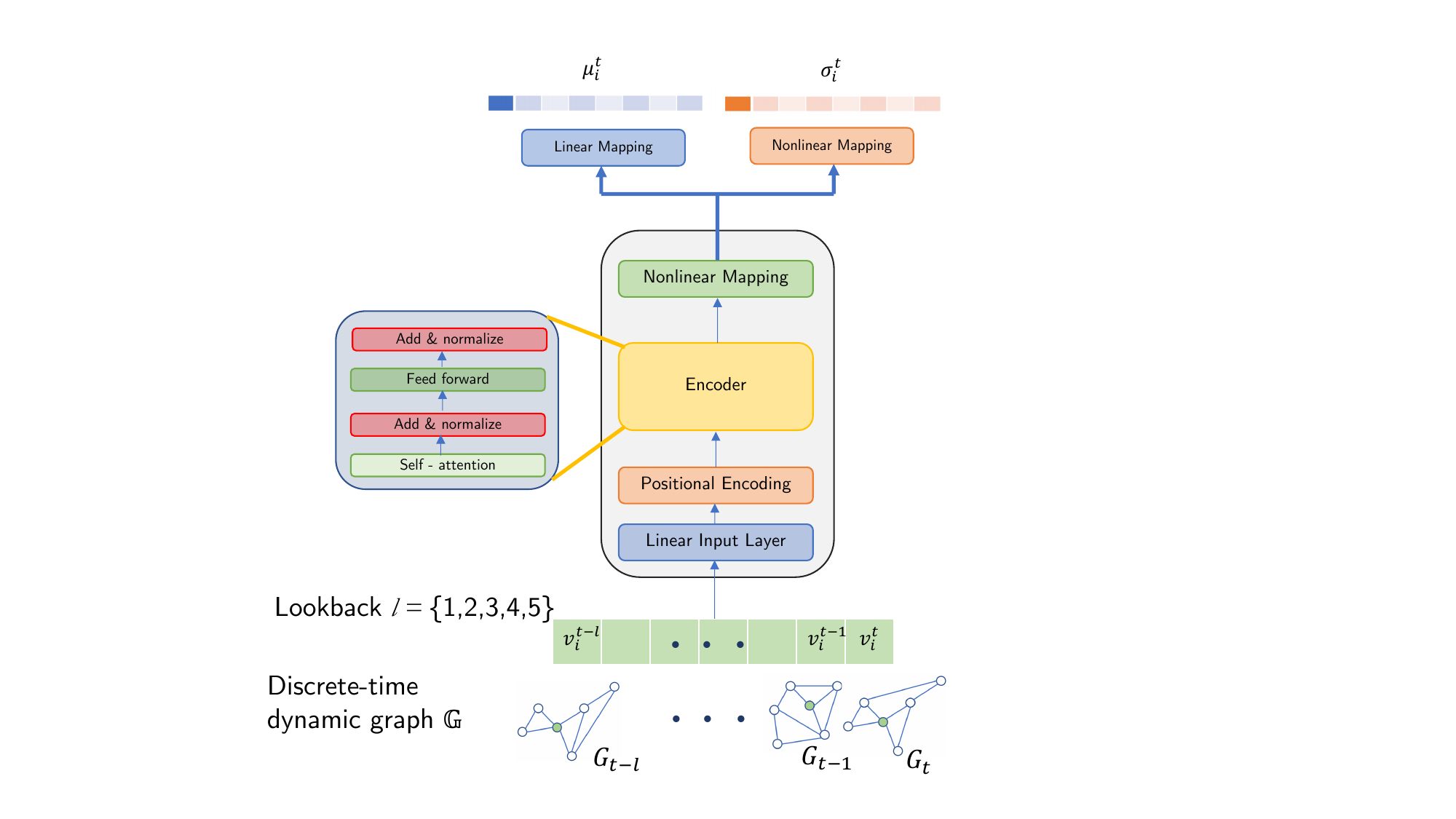}
    \caption{Illustration of our proposed TransformerG2G model architecture. The attention mechanism of the transformer enables the model to capture long-term context in the dynamics of the graph. The model's input is the history of a node over $l + 1$ steps, from timestamps $t-l$ to $t$ and outputs the embedding vectors at timestamp $t$ of that particular node. Here, the lookback $l$ is a hyperparameter and different values of $l$ are considered in our experiments.}
    \label{fig:transformerG2G}
\end{figure}

During preprocessing, the original adjacency matrices $A_t$, with shape $(n_t, n_t)$, are modified to obtain the modified adjacency matrix $\tilde{A}_t$, ensuring that all the modified adjacency matrices have the same shape of $(n, n)$. Here $n = \lvert \cup_{t=1}^{T}{V_t}\rvert\ $ is the maximum number of nodes across all timestamps, and if a node is absent from the graph snapshot during a particular timestamp, the row and column in $\tilde{A}$ corresponding to that node is filled with zeroes, indicating that node is not connected with any of the other nodes. The TransformerG2G model takes as input the node history of a node \textit{i} represented by a sequence of vectors $(v_i^{t-l}, \cdots, v_i^{t-1}, v_i^t)$, where $v_i^t \in \mathbb{R}^n $ and $v_i^t$ corresponds to the $i$th row of $\tilde{A}_t$. Every vector in this input sequence is first projected to a $d$-dimensional space using a linear layer, followed by adding a vanilla positional encoding to it. In our work, we use the same positional encoding as in \citep{vaswani2017attention}. The resulting sequence of $d$-dimensional vectors is then processed by an encoder block, which outputs another sequence of $d$-dimensional vectors. The encoder block used has the same architecture as the original transformer paper \cite{vaswani2017attention}, and consists of i) a self-attention layer, ii) a residual connection followed by layer normalization (add \& norm), iii) a feed-forward network, and again a iv) residual connection followed by layer normalization (add \& norm).  The sequence of output vectors from the encoder block is concatenated and passed through a linear layer followed by $tanh$ activation to obtain a single vector $h_i$. This vector is then fed through two projection heads to obtain the embeddings $\mu_i^t \in \mathbb{R}^{L_o}$ and $\sigma_i^t \in \mathbb{R}^{L_o}$, representing the node $v_i^t$. The projection head to obtain the mean vector consists of a linear layer, i.e., $\mu_i = h_iW_{\mu} + b_{\mu}$, and the projection head to obtain the covariance vector consists of a linear layer with $elu$ activation, i.e., $\sigma_i = elu(h_i W_{\Sigma} + b_{\Sigma}) + 1,$ to ensure the positivity of the covariance vector. The self-attention mechanism in the encoder block enables the model to learn the information from the node's historical context while predicting the node's embedding. Specifically, we employ the scaled dot-product attention mechanism \cite{vaswani2017attention} in our TransformerG2G model, defined as follows.

\begin{equation}
    Attention(Q,K,V) = softmax\left(\frac{QK^T}{\sqrt{d}}\right)V,
\end{equation}
where $Q$ denotes the query matrix, $K$ denotes the key matrix and $V$ denotes the value matrix. The $Q$, $K$ and $V$ matrices are obtained by stacking the output sequence of the positional encoding layer to form a matrix $Z$ of shape $(l+1,d)$, and then passing it through three separate linear layers, i.e., $Q = ZW_Q$, $K = Z W_K$ and $V = Z W_V$. The attention mechanism plays a crucial role in capturing long-term dependencies and modeling relationships between different timestamps in the graph. Next, we discuss the loss function and the training methodology of the TransformerG2G model.

The TransformerG2G model is trained using a triplet-based contrastive loss \cite{bojchevski2017deep} defined in Eq.~\ref{eq:triplet_loss}. With each node as reference, we first sample the $k$-hop neighborhoods, that is, nodes that are exactly $k$ hops away from a specific node. We then use these $k$-hop neighborhoods to sample a node triplet set $\mathcal{T}_t = \{ (v_i, v_i^{near}, v_i^{far})| v_i \in V_t\}$, where $v_i$ is the reference node, $v_i^{near}$ is a node that is close to $v_i$, and $v_i^{far}$ is a node far from the reference node. The triplet satisfies the constraint that the shortest path between the reference node $v_i$ and $v_i^{near}$ is smaller than the shortest path between the reference node $v_i$ and  $v_i^{far}$, i.e., $sp(v_i, v_i^{near}) < sp(v_i, v_i^{far})$. Here $sp( )$, indicates the length of the shortest path between two nodes. The triplet-based contrastive loss is as follows:
\begin{equation}
\label{eq:triplet_loss}
    \mathcal{L} = \sum_t \sum_{\left(v_i, v_i^{near}, v_i^{far}\right) \in \mathcal{T}_t} \left[ \mathbb{E}^2_{\left(v_i, v_i^{near}\right)}  + e^{-\mathbb{E}_{\left(v_i, v_i^{far}\right)}}\right],
\end{equation}
where $\mathbb{E}_{\left(v_i, v_i^{near}\right)}$ and $\mathbb{E}_{\left(v_i, v_i^{far}\right)}$ Kullback-Leibler divergence (KL divergence) between the embeddings of reference node and near-by node, and the embeddings of reference node and far-away node, respectively. Here, KL divergence is used as a metric for measuring the dissimilarity between the joint normal distributions of two nodes in the embedding space. The specific formula of the KL divergence between the multivariate Gaussian embeddings of two nodes ($v_i$ and $v_j$) is shown in Eq.~\ref{eq: kl_divergence}.
\begin{equation}
\label{eq: kl_divergence}    
\begin{split}
    \mathbb{E}_{\left( v_i, v_j \right)} &= D_{KL}\left( \mathcal{N}(\mu_i,\Sigma_i), \| \mathcal{N}(\mu_j, \Sigma_j)  \right)\\
    &= \frac{1}{2} \left[ tr(\Sigma_j^{-1} \Sigma_i) + (\mu_j - \mu_i)^T \Sigma_j^{-1}(\mu_j-\mu_i) - L + log\frac{|\Sigma_j|}{|\Sigma_i|}\right].
\end{split}
\end{equation}

By employing the TransformerG2G model, we aim to obtain lower-dimensional multivariate Gaussian representations of nodes, that effectively capture long-term temporal dynamics with varying lengths of temporal node context.

%%%%%%%%%%%%%%%%%%%%%%%%%%%%%%%%%%%%%%%%%%%%%
\section{Experiments} \label{experiments}
%%%%%%%%%%%%%%%%%%%%%%%%%%%%%%%%%%%%%%%%%%%%%
\subsection{Dataset descriptions}
\label{sec:dataset}

We evaluated our proposed TransformerG2G model on six different dynamic graph benchmarks with varying evolutionary dynamics. The specific graph dataset descriptions are shown in Table~\ref{tab:benchmark_table}. 
\begin{table}[ht]
\centering
\caption{Details of the experimental datasets.}
\label{tab:benchmark_table}
\begin{tabular}{l||lllll}
     \hline
     Dataset & \#Nodes & \#Edges & \#Timestamps   & \#Train/Val/Test & Embedding Size ($L_o$)\\ 
     \hline \hline
     SBM & 1,000 & 4,870,863 & 50 & 35/5/10 & 64\\
     Reality Mining & 96 & 1,086,403 & 90 & 63/9/18 & 64\\
     UCI & 1,899 & 59,835 & 88 & 62/9/17 & 256\\
     Slashdot & 50,825 & 42,968 & 12 & 8/2/2 & 64\\
     Bitcoin-OTC & 5,881 & 35,588 & 137 & 95/14/28 & 256\\
     AS & 65,535 & 13,895 & 100 & 70/10/20 & 64\\
     \hline
\end{tabular}
\end{table}

% synthetic dataset
\textbf{Stochastic Block Model (SBM) dataset\footnote{\url{https://github.com/IBM/EvolveGCN/tree/master/data}}}: This dataset is generated by using the Stochastic Block Model (SBM) model. Here, the first snapshot is generated so that there are three equal-sized communities with in-block probability 0.2 and a cross-block probability of 0.01. Subsequent graph snapshots are generated by randomly picking 10-20 nodes at each time instant and moving them to a different community. The SBM dataset consists of 1000 nodes, 4,870,863 edges and 50 timestamps.

% social network
\textbf{Reality Mining dataset\footnote{\url{http://realitycommons.media.mit.edu/realitymining.html}}}: The Reality Mining dataset is a human contact network among 100 students of the Massachusetts Institute of Technology (MIT). The data was collected using 100 mobile phones over a span of 9 months in 2004. In the graph snapshots, each node represents a student and an edge indicates physical contact between two nodes. In our experiment, the dataset contains 96 nodes and 1,086,403 undirected edges across 90 timestamps.

% social network
\textbf{UC Irvine messages (UCI) dataset\footnote{\url{http://konect.cc/networks/opsahl-ucsocial/}}}: This dataset contains messages sent between the users of the online student community at the University of California, Irvine. Within the UCI dataset, there are 1,899 nodes and 59,835 edges across 88 timestamps, forming a directed graph. This dataset exhibits highly transient dynamics.

% social network
\textbf{Slashdot dataset\footnote{\url{http://konect.cc/networks/slashdot-threads/}}}: It is a large-scale social reply network for the technology website Slashdot. Within this network, the nodes represent users and the edges correspond to the replies of users. The edges are directed and originate from the responding user. Edges also have the timestamp of the reply. The Slashdot dataset contains 50,824 nodes and 42,968 edges over 12 timestamps.

% finance network 
\textbf{Bitcoin-OTC (Bit-OTC) dataset\footnote{\url{http://snap.stanford.edu/data/soc-sign-bitcoin-otc.html}}}: It is who-trusts-whom network of people who trade using Bitcoin on a platform called Bitcoin OTC. The Bit-OTC dataset has 5,881 nodes and 35,588 edges with 137 timestamps, forming a directed graph. This dataset displays highly transient dynamics.

% social network
\textbf{Autonomous Systems (AS) dataset\footnote{\url{https://snap.stanford.edu/data/as-allstats.html}}}: It consists of a communication network of who-talks-to-whom from the BGP (Border Gateway Protocol) logs. The AS dataset used in our experiment has 65,535 nodes and 13,895 edges with 100 timestamps in total.

%%%%%%%%%%%%%%%%%%%%%%%%%%%%%%%%%%%%%%
\subsection{Temporal edge appearance (TEA) plot}
\label{sec:TEAplot}

\begin{figure}[p]
    \centering
    \includegraphics[width = .94\textwidth]{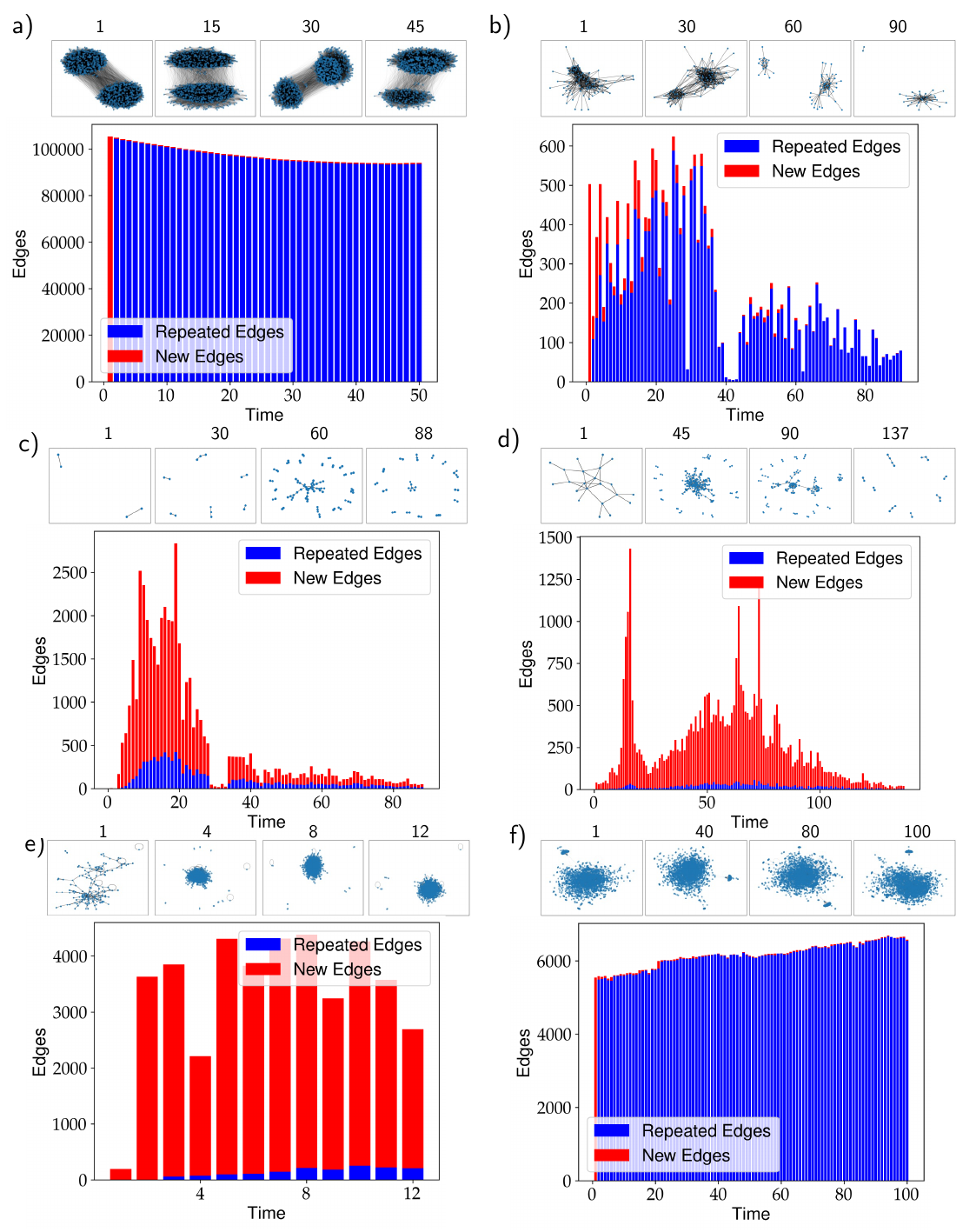}
    \caption{Six different discrete-time temporal graph benchmarks along with their corresponding Temporal Edge Appearance (TEA) plots for effective characterization of evolutionary dynamics. a)-f) correspond to SBM, Reality Mining, UCI, Bitcoin, Slashdot, and Autonomous Systems, respectively. For each benchmark, the top row displays four distinct graph snapshots with the timestamp numbers, and the bottom row shows the TEA plot including the number of newly added edges (red bars) and the number of repeated edges (blue bars) over time. The $novelty$ index values corresponding to a)-f) are 0.0252, 0.0761, 0.7526, 0.9161, 0.9861, and 0.014, respectively.}
    \label{fig:TEA_graphsnapshots}
\end{figure}

To characterize the temporal dynamic patterns in dynamic graphs, we utilize the temporal edge appearance (TEA) plot proposed in~\cite{poursafaei2022towards} to effectively quantify the proportion of repeated edges compared to newly observed edges at each timestamp. Moreover, the average ratio of new edges in each timestamp can be calculated as the $novelty$ index for every benchmark using Eq.~\ref{eq:TEA_formula}.
\begin{equation}
novelty  = \frac{1}{T}\sum_{t=1}^T \frac{|E_t \backslash E_{seen}^t|}{|E_t|}, 
\label{eq:TEA_formula}    
\end{equation}
where $E_t$ represents a set of edges at timestamp $t$, $E_{seen}^t$ denotes the set of edges seen in the previous timestamps $\{1, \cdots, t-1\}$. 

The top row in each subfigure presents visualizations of four graph snapshots selected from arbitrary timestamps, and the bottom row shows the corresponding temporal edge appearance (TEA) plot for characterizing the temporal dynamics as described in the Section~\ref{sec:TEAplot}.

In Fig.~\ref{fig:TEA_graphsnapshots}, the bottom row in each subfigure presents the corresponding TEA plot, where the blue bars represent the number of repeated edges, and the red bars denote the number of newly added edges. The $novelty$ index values computed for the six dynamic graph benchmarks in a)-f) are 0.0252, 0.0761, 0.7526, 0.9161, 0.9861, and 0.014, respectively. Specifically, Fig.~\ref{fig:TEA_graphsnapshots}a) and f) with small $novelty$ index values present relatively smooth changes in dynamics over time, while Fig.~\ref{fig:TEA_graphsnapshots}b) presents medium changing dynamics with more graph topological changes in the first 20 snapshots but smaller changes for the last 20 snapshots. In contrast, Fig.~\ref{fig:TEA_graphsnapshots}c)-e) achieves the highest $novelty$ values and hence it exhibits highly evolutionary dynamics with a significant number of newly added edges over different timestamps. Clearly, the benchmark datasets comprises of datasets with very rich and unique evolutionary dynamics, and hence, including the historical context contributes to learning more accurate graph embeddings.

%%%%%%%%%%%%%%%%%%%%%%%%%%%%%%%%%%%%
\subsection{Cosine similarity}
\label{sec:cosine_similarity}
% cosine similarity plot
\begin{figure}[!ht]
    \centering
    \includegraphics[width = \textwidth]{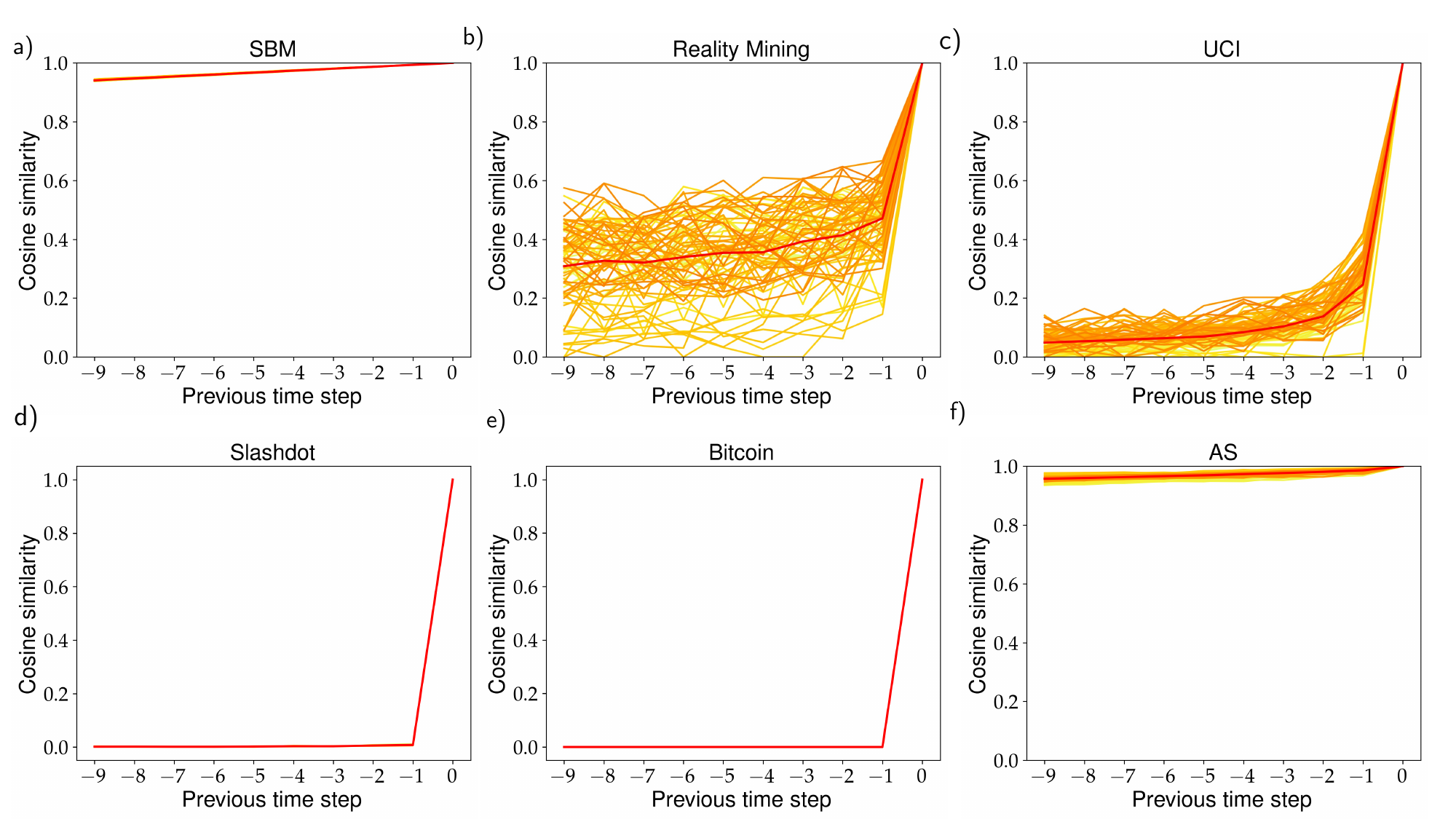}
    \caption{Cosine similarity plots with previous timestamps for six dynamic graph datasets. a)-f) correspond to the SBM, Reality Mining, UCI, Bitcoin, Slashdot, and Autonomous Systems datasets. The red curve indicates the average cosine similarity.}
    \label{fig: cosine_similarity_plot}
\end{figure}
In order to measure the temporal graph correlations between the current graph snapshot ($t$) and its preceding ones ($t' = t, \cdots, t-9)$ over time, we compute the cosine similarity between graph snapshot at timestamp $t$ ($t = 10, 11, \cdots, T$) and its preceding ones over the previous 10 timestamps. To compute the cosine similarity between two snapshots at time $t$ and $t'$, we first modify the original adjacency matrices $A_t$, with shape $(n_t, n_t)$, to obtain the modified adjacency matrices $\tilde{A}_t$, ensuring that they have the same shape of $(n, n)$. Now, the similarity between graph snapshots at $t$ and $t'$ is computed by calculating the cosine similarity between the flattened modified adjacency matrices, $\tilde{A}_t$ and $\tilde{A}_{t'}$. The corresponding results for all six dynamic graph benchmarks are shown in Fig.~\ref{fig: cosine_similarity_plot}. The red curve in each figure represents the average cosine similarity over all curves corresponding to different time windows (window size = 10). We can observe that the SBM and AS datasets (with low $novelty$) exhibit subtle changes in graph structures over time, as indicated by the stable high cosine similarities between the current timestamp and the previous 10 timestamps. However, the UCI, Slashdot and Bitcoin datasets (with high $novelty$) show a sharp drop in the cosine similarity with preceding timestamps, indicating more substantial changes in the graph structures over time compared to the SBM and AS datasets. Fig.~\ref{fig: cosine_similarity_plot} demonstrates varying temporal evolving patterns across different timestamps for different benchmarks, hence highlighting the importance of incorporating historical context to capture temporal dynamics of graphs.

%%%%%%%%%%%%%%%%%%%%%%%%%%%%%%%%%%%%%
\subsection{Implementation details}

Evaluating graph embedding algorithms involves two stages: i) model training and obtaining embeddings for every node for each timestamp, ii) evaluating the performance on a downstream task, such as link prediction and node classification. In our experiments, downstream link prediction task is performed to demonstrate the goodness of the learned embeddings. The accuracy metrics on the link prediction task are used as a proxy to quantify the performance of the embedding algorithm. For training both the TransformerG2G model and the classifier, we utilize the first 70\% of the timestamps for training, the next 10\% of timestamps for validation and the remaining 20\% of the timestamps for testing. The number of train/val/test timestamps for each dataset is shown in Table \ref{tab:benchmark_table}. 

\textbf{Multistep method:}
The multistep method was trained using the same hyperparameters as the DynG2G paper \cite{xu2022dyng2g}. The input vectors to the linear layer is an $n$-dimensional vector. The linear layer followed by the $ReLU$ activation function, results in a 512 dimensional hidden vector $h$. This hidden vector is mapped using two projection heads to get the mean vector $\mu_i$ and the diagonal covariance vector $\sigma_i$, both of dimension $L_o$.

\textbf{TransformerG2G:} The dimensionality $n$ of the vectors in the input sequence to TransformerG2G is the maximum number of nodes in the graph dataset and is shown in Table \ref{tab:benchmark_table} for all the benchmark datasets. The first linear layer projects these input vectors to a $d$-dimensional space. We use $d = 256$ for SBM, Reality Mining, Slashdot and Autonomous System datasets, and $d=512$  for UCI and Bitcoin datasets. In the transformer, we used one encoder layer with a single attention head. In the nonlinear layer after the encoder, a $tanh$ activation function and the vectors were projected to a $512$ dimensional space, similar to the DynG2G paper. In the nonlinear projection head we used an $elu$ activation function. To optimize the weights of the TransformerG2G model, we used the Adam optimizer for all datasets. The learning rate was set to 1e-4 for SBM, Reality Mining and Autonomous Systems. In the case of UCI, Slashdot and Bitcoin we used a learning rate of 1e-6. The first 70\% of the timestamps was used for training the TransformerG2G model, and the next 10\% of the timestamps was used for validation. Once the model was trained, we used the trained model to predict and save the embeddings of the nodes for all the timestamps. Please note that we have a single model for all the timestamps, whereas DynG2G had a different model for each training timestamp.

\textbf{Link prediction task:} In the downstream link prediction task, we used a classifier to predict whether two nodes have a direct link or not, given the embeddings of the two nodes. The classifier was a MLP with one hidden layer with $L_o$ neurons in the hidden layer and  $ReLU$ activation. The classifier takes as input the embeddings ($\mu_i$ and $\mu_j$) of two nodes.   The classifier was trained to minimize the weighted binary cross-entropy loss using an Adam optimizer with a learning rate of 1e-4. For this task as well, the data is split into train/val/test as shown in Table \ref{tab:benchmark_table}. The classifier was trained on the first 70\% of the timestamps, and is used to predict the links on the testing timestamps. The average MAP and MRR values over the graphs during the testing timestamps are reported to quantify the goodness of the embeddings.

We use two evaluation metrics to quantify the performance on the downstream link prediction taks: i) the mean average precision (MAP) and ii) the mean reciprocal rank (MRR). The definitions of MAP and MRR are as follows\cite{xu2022dyng2g}: \\
\textbf{a) Mean average precision (MAP)}: Here we rank all the predictions of node $q$ in the descending order of their probabilities. The average precision ($AP$) is then computed for each node using the formula specified in Eq.~\ref{eq:MAP_definitions}. Here, $m$ represents the number of connections predicted by our model for a specified node $q$,  $n$ represents the maximum number of connections to be considered, $P(k)$ denotes the precision at $k$-th node, and $rel(k)$ is equal to 1 if $k$-th node is connected to node $q$, otherwise it is 0. The $MAP$ value is obtained by averaging the $AP$ values of all the nodes $Q$ (see Eq.~\ref{eq:MAP_definitions}). 
\begin{equation}
\centering
MAP = \frac{1}{|Q|}\sum_{q \in Q}{AP(q)}; \,\, AP(q) =\frac{1}{m} \sum_{k=1}^n{[P(k)\times rel(k)]}.
\label{eq:MAP_definitions}
\end{equation}
\textbf{b) Mean reciprocal rank (MRR)}. $MRR$ is calculated by ranking all the predictions of node $q$ in descending order of probabilities. The reciprocal rank ($1/k_q$) is then computed by determining the rank position ($k_q$) of the first relevant linked node for the query node $q$. The MRR value is obtained by taking the average reciprocal rank of all the nodes. The detailed formula for computing MRR is specified in Eq. \ref{eq:MRR_definitions}. It is important to note that while MRR considers only the highest-ranked relevant node, MAP takes into account the ranking of all relevant nodes. MRR therefore provides a general measure of the quality for link prediction, wheras MAP considers the ranking of all relevant nodes. \begin{equation}
\centering
\label{eq:MRR_definitions}
MRR = \frac{1}{|Q|}\sum_{q\in Q}{\frac{1}{k_q}}
\end{equation}
% \textcolor{red}{\textbf{Mengjia: }Somewhere we need to illustrate what is link prediction?  
% how to learn the learned Gaussian embedding with transformers including train, val and test the model. 
% Moreover, we need to present the details of link prediction by training a new MLP classifiers and then apply it to predict the links, i.e., output link scores.} Done

%%%%%%%%%%%%%%%%%%%%%%%%%%%%%%%%%
\subsection{Results of multi-step method}
\begin{figure}[ht]
    \centering
    \includegraphics[width = \textwidth]{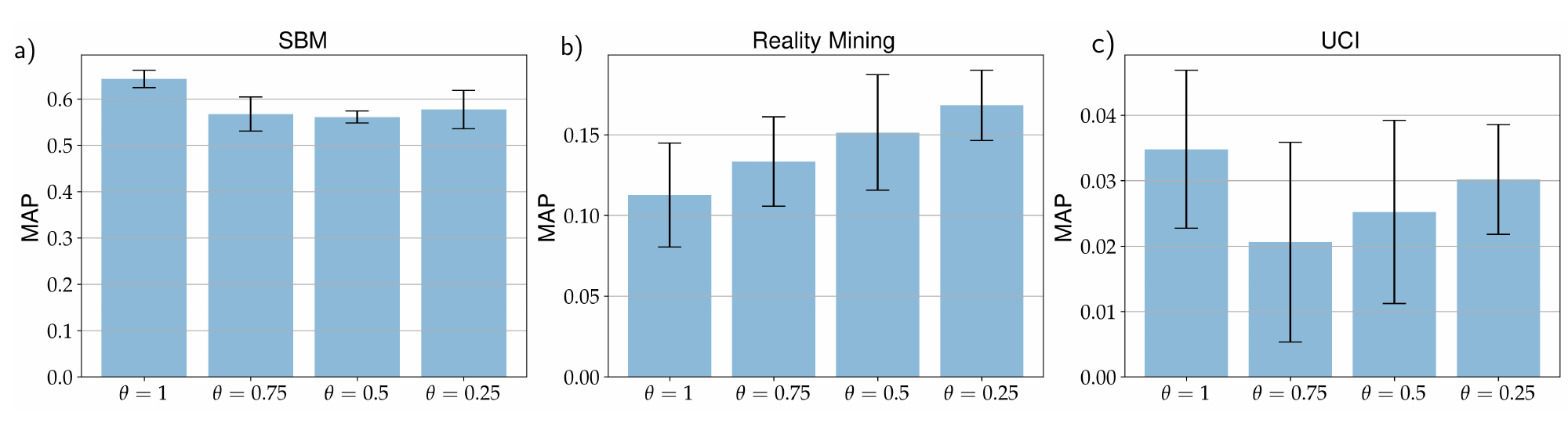}
    \caption{MAP value for link prediction using two-step method with SBM, Reality Mining and UCI datasets.}
    \label{fig: two_theta_plot}
\end{figure}

For the two-step method, we tested different $\theta$ parameters to initialize the weight matrix at timestamp $t$ by transferring the weights learned from the previous two timestamps $t-1 $ and $t-2$ with coefficients determined by $\theta$ and $1-\theta$. Fig.~\ref{fig: two_theta_plot} shows three bar plots illustrating the mean average precision (MAP) values (see its definition in our prior work~\cite{xu2022dyng2g}) for temporal link prediction using SBM, Reality Mining and UCI datasets. For the SBM dataset, varying $\theta$ does not result in a significant improvement in MAP or MRR. Since SBM is a synthetic dataset with Markovian behavior, i.e., the graph at the next timestamp depends only on the current timestamp, including more history does not enhance the quality of the learned embeddings. In the case of Reality Mining, we see that it achieves higher MAP and MRR values using the two-step method compared to the ones using DynG2G ($\theta = 1$). An intriguing observation from the temporal link prediction results in Fig.~\ref{fig: two_theta_plot} b) is that the MAP value increases as $\theta$ decreases (i.e., placing more emphasis on $t-2$). In the case of UCI dataset, we do not see an improvement in the MAP or MRR by using the two-step method despite of the fact that the data are non-Markovian, indicating the necessity of using more complex models.

Additionally, Table \ref{tab:twosteps_results} provides a comprehensive overview of the numerical results for the temporal link prediction task. The table includes the MAP and mean reciprocal rank (MRR) values achieved by utilizing stochastic graph embeddings learned from the modified DynG2G model with two-step approach. 
%%%%%%%%%%%%%%%%%%%%%%%%
\begin{table}[ht]
\centering
\caption{Link prediction performance based on the two-step method in terms of MAP and MRR (bold text indicates the best performance achieved for each benchmark).}
\vspace{.2cm}
\label{tab:twosteps_results}
\begin{tabular}{c||c||cccc}
\hline
Benchmark & Metric  & $\theta = 1$  & $\theta = 0.75$ & $\theta = 0.5$   & $\theta = 0.25$ \\ \hline\hline
% SBM
\multicolumn{1}{c||}{\multirow{2}{*}{SBM}} & MAP & \textbf{0.6433 $\pm$ 0.0187} & 0.5671 ± 0.0371 &  0.5609 ± 0.0127 & 0.5772 ± 0.0413\\ %\cline{2-6}
\multicolumn{1}{c||}{}                     & MRR & \textbf{0.0409 $\pm$ 0.0014} &0.0344 ± 0.0034 &0.0359 ± 0.0025 & 0.0353 ± 0.0027\\ \hline
% Reality Mining
\multicolumn{1}{c||}{\multirow{2}{*}{Reality Mining}} & MAP & 0.1126 ± 0.0322 & 0.1334 ± 0.0277 &  0.1514 ± 0.0358 & \textbf{0.1682 ± 0.0218}\\ %\cline{2-6}
\multicolumn{1}{c||}{}                     & MRR & 0.1973 ± 0.0341 & 0.2027 ± 0.0402 & \textbf{0.2445 ± 0.0286} & 0.2289 ± 0.0234\\\hline
% UCI
\multicolumn{1}{c||}{\multirow{2}{*}{UCI}} & MAP & \textbf{0.0348 ± 0.0121} & 0.0206± 0.0153 & 0.0252 ± 0.0140 & 0.0302 ± 0.0084\\ %\cline{2-6}
\multicolumn{1}{c||}{}                     & MRR & \textbf{0.3247 ± 0.0312} & 0.1825 ± 0.0416 & 0.2349 ± 0.0502 & 0.2674 ± 0.0424\\\hline
\end{tabular}
\end{table}
%%%%%%%%%%%%%%%%%%%%%%%%%%%%%%
To further improve the expressivity of the DynG2G model~\cite{xu2022dyng2g}, we consider a ``three-step method'' that enables capturing heterogeneous dynamics learned from the previous three timestamps. The specific weight initialization schema is presented in Eq.~\ref{eq:three-step-method}. The weight matrix at timestamp $t$ is initialized by transferring the learned weight matrices from three previous timestamps with different coefficients ($\theta_1, \theta_2, \theta_3$), where $\theta_1 + \theta_2 + \theta_3 = 1$.
%  three-step method
\begin{equation}
    W_{t} = \theta_1 W_{t-1} + \theta_2 W_{t-2} + \theta_3 W_{t-3}, \quad \text{where} \quad \theta_1 + \theta_2 + \theta_3 = 1.
\label{eq:three-step-method}
\end{equation}

The corresponding temporal link prediction results on the UCI dataset with three-step approach are shown in Table.~\ref{tab:3steps_results}. The experimental results show that by leveraging more temporal node context, the three-step method achieves a higher MAP value compared to the two-step method on the UCI dataset. However, we note that the three-step method results in a lower MRR than the two-step method. Although, in the context of node embedding and classification MAP is often preferred over MRR, the lower MRR points out the difficulty of manually integrating historical data using a multi-step method. Hence, utilizing a self-attention-based architecture is a more practical choice for accommodating the historical dependencies, and this hypothesis is also supported by the results.

\begin{table}[ht]
\centering
\caption{Link prediction performance based on three-step method in terms of MAP and MRR (bold text indicates the best performance achieved for the benchmark).}
\vspace{.2cm}
\label{tab:3steps_results}
\begin{tabular}{c||c||ccc|ccc|ccc|ccc}
\hline
\multicolumn{1}{c||}{\multirow{2}{*}{Benchmark}} & \multicolumn{1}{|c||}{\multirow{2}{*}{Metric}}  & $\theta_{1}$ & $\theta_{2}$ & $\theta_{3}$ & $\theta_{1}$ & $\theta_{2}$ &  $\theta_{3}$ & $\theta_{1}$ & $\theta_{2}$ &  $\theta_{3}$ \\ \cline{3-11}

 &   & 0.2 & 0.2 & 0.6 & 0.6 & 0.2 &0.2 & 0.2 &0.6 & 0.2 \\ \hline\hline
% UCI
\multicolumn{1}{c||}{\multirow{2}{*}{UCI}} & MAP & \multicolumn{3}{|c|}{0.0355 ± 0.0154} & \multicolumn{3}{|c|}{0.0341± 0.0064} & \multicolumn{3}{|c|}{\textbf{0.0406 ± 0.0215}} \\  %\cline{2-11}
\multicolumn{1}{c||}{}                     & MRR & \multicolumn{3}{|c|}{0.1840 ± 0.0298} & \multicolumn{3}{|c|}{\textbf{0.2074 ± 0.0080}} & \multicolumn{3}{|c|}{0.1941 ± 0.0304} \\\hline
\end{tabular}
\end{table}

%%%%%%%%%%%%%%%%%%%%%%%%%%%%%%%%%%
\subsection{Results of TransformerG2G model}
\begin{figure}[ht]
    \centering
    \includegraphics[width = \textwidth]{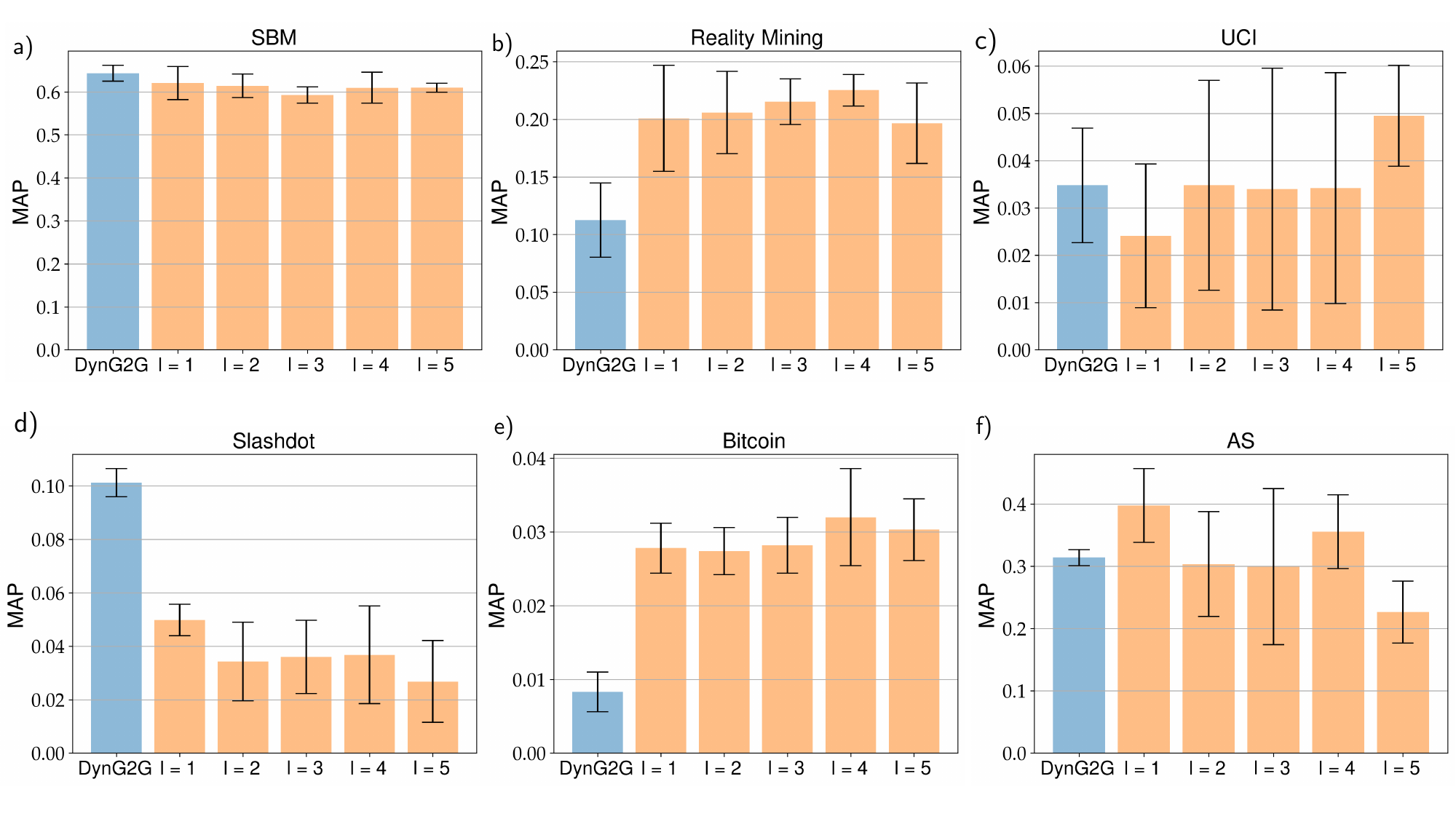}
    \caption{MAP values for temporal link prediction based on node embeddings learned with our proposed TransformerG2G model across six dynamic graph datasets, where higher MAP values indicate better link prediction performance. The datasets corresponding to a)-f) are SBM, Reality Mining, UCI, Bitcoin, Slashdot, and Autonomous Systems, respectively. The blue bars display the MAP values achieved by DynG2G, while the yellow bars present the MAP values obtained by our TransformerG2G model with different lookbacks ($l = \{1,2,3,4,5\}$).}
    \label{fig: MAP_plot}
\end{figure}

\begin{table}[ht]
\centering
\caption{Link prediction performance based on TransformerG2G method in terms of MAP and MRR for different lookbacks ($l$). For each experiment, the code was run 5 times, and the mean and standard deviation are reported.}
\vspace{.2cm}
\label{tab:tranformerg2g_results}
\scalebox{0.85}{
\begin{tabular}{ c || c c c c c c }
\hline
Benchmark & Metric  & $l = 1$  & $l = 2$ & $l = 3$   & $l = 4$ & $l = 5$ \\
\hline
\hline
% SBM
\multicolumn{1}{ c ||}{\multirow{2}{*}{SBM}} & MAP & \textbf{0.6204 $\pm$ 0.0386} & 0.6143 ± 0.0274 &  0.5927 ± 0.0192 & 0.6096 ± 0.0360 & 0.6097 ± 0.0104\\ 
\multicolumn{1}{ c ||}{}                     & MRR & 0.0369 $\pm$ 0.0055 & 0.0388 ± 0.0009 &0.0317 ± 0.0053 & 0.0349 ± 0.0059 & 0.0386 ± 0.0002\\\hline
% Reality Mining
\multicolumn{1}{ c ||}{\multirow{2}{*}{Reality Mining}} & MAP & 0.2010 ± 0.0460 & 0.2057 ± 0.0356 &  0.2153 ± 0.0198 & \textbf{0.2252 ± 0.0137} & 0.1965 ± 0.0349\\
\multicolumn{1}{ c ||}{}                     & MRR & 0.1414 ± 0.0046 & 0.1306 ± 0.0094 & 0.1334 ± 0.0094 & 0.1294 ± 0.0129 & 0.1311 ± 0.0083\\\hline
% UCI
\multicolumn{1}{ c ||}{\multirow{2}{*}{UCI}} & MAP & 0.0241 ± 0.0152 & 0.0347± 0.0222 & 0.0340 ± 0.0256 & 0.0342 ± 0.0244 & \textbf{0.0495 ± 0.0107}\\ 
\multicolumn{1}{ c ||}{}                     & MRR & 0.2616 ± 0.0879 & 0.2947 ± 0.1087 & 0.2905 ± 0.1121 & 0.3042 ± 0.1115 & 0.3447 ± 0.0196\\\hline
% Slashdot
\multicolumn{1}{ c ||}{\multirow{2}{*}{Slashdot}} & MAP & \textbf{0.0498 ± 0.0059} & 0.0343± 0.0147 & 0.0360 ± 0.0137 & 0.0368 ± 0.0183 & 0.0268 ± 0.0153\\ 
\multicolumn{1}{ c ||}{}                     & MRR & 0.2568 ± 0.0126 & 0.2679 ± 0.0222 & 0.2824 ± 0.0143 & 0.2783 ± 0.0358 & 0.2749 ± 0.0096 \\\hline
% Bitcoin
\multicolumn{1}{ c ||}{\multirow{2}{*}{Bitcoin}} & MAP & 0.0278 ± 0.0034 & 0.0274± 0.0032 & 0.0282 ± 0.0038 & 0.0319 ± 0.0066 & \textbf{0.0303 ± 0.0042}\\ 
\multicolumn{1}{ c ||}{}                     & MRR & 0.3194 ± 0.0182 & 0.3415± 0.0246 & 0.3788 ± 0.0292 & 0.4089 ± 0.0079 & 0.4037 ± 0.0188\\\hline
% AS
\multicolumn{1}{ c ||}{\multirow{2}{*}{AS}} & MAP & \textbf{0.3976 ± 0.0594} & 0.3035 ± 0.0842 & 0.2994 ± 0.1252 & 0.3557 ± 0.0592 & 0.2266 ± 0.0498\\ 
\multicolumn{1}{ c ||}{}                     & MRR & 0.3599 ± 0.0276 & 0.3412 ± 0.0162 & 0.3468 ± 0.0218 & 0.3550 ± 0.0215 & 0.3319 ± 0.0203\\\hline

\end{tabular}}
\end{table}
In Table~\ref{tab:tranformerg2g_results} we show the results for temporal link prediction on the six benchmark datasets, using embeddings obtained from our TransformerG2G model for different lookbacks $l$. We also visualize the MAP values using bar plots for temporal link prediction across these six different benchmarks in Fig.~\ref{fig: MAP_plot}. In Fig.~\ref{fig: MAP_plot}, the blue bars shows the MAP using the embeddings achieved by the DynG2G model, and the orange bars shows the MAP using the embeddings achieved by our proposed TransformerG2G method. In the case of SBM dataset, we see that varying the lookback changes the MAP only slightly. This behaviour is expected since SBM is a Markovian dataset, and therefore including more history would not improve the MAP and MRR. In the case of Reality Mining, we see an improvement in MAP till $l = 4$. The MAP is better than that of DynG2G for all lookbacks. In the case of UCI dataset, the mean MAP at $l = 5$ is larger than that for DynG2G. We also see a large variance in the MAP for UCI, possibly due to the rapid change in the number of nodes in this dataset over time. In the case of Slashdot, the MAPs using the TransformerG2G model is lower than that for the DynG2G model for all the lookbacks. This is due to the very high novelty in this dataset, which indicates that Slashdot has highly evolutionary dynamics and therefore it is difficult for the TransformerG2G model to accurately learn the dynamics. In the case of Bitcoin, the MAP values using the TransformerG2G model is higher than that using the DynG2G model for all lookbacks. The novelty of Bitcoin is similar to that of Slashdot, however, Bitcoin has 137 timestamps compared to Slashdot with 12 timestamps. The larger number of timestamps translates to more data, allowing the TransformerG2G model to learn the patterns in the data despite having a high novelty. In the case of AS dataset as well, we were able to attain higher MAP using the TransformerG2G model.

%%%%%%%%%%%%%%%%%%%%%%%%%%%%%%%%%%
\subsection{Comparison of TransformerG2G with other baselines}

In the Table \ref{tab:method_comparisons_LP} we show the comparison of the temporal link prediction task against other embedding algorithms for the six benchmarks.

To evaluate the performance of our model, TransformerG2G, in the task of temporal link prediction, we compare the proposed TransformerG2G with other baselines: DynGEM~\cite{goyal2018dyngem}, dyngraph2vecAE~\cite{goyal2020dyngraph2vec}, dyngraph2vecAERNN \cite{goyal2020dyngraph2vec}, EvolveGCN~\cite{pareja2020evolvegcn}, ROLAND~\cite{you2022roland} and DynG2G~\cite{xu2022dyng2g}. Brief descriptions about these methods are presented below.
\begin{itemize}
    \item \textit{DynGEM~\cite{goyal2018dyngem}}: This is a dynamic graph embedding method that uses autoencoders to obtain graph embeddings. The weights of the previous timestamp are used to initialize the weights of the current timestamp, and thereby makes the training process faster.
    \item \textit{dyngraph2vecAE~\cite{goyal2020dyngraph2vec}}: This method uses temporal information in dynamic graphs to obtain embeddings using an autoencoder. The encoder takes in the historical information of the graph, and the decoder reconstructs the graph at the next timestamp, and the latent vector corresponds to the embedding of the current timestamp.
    \item \textit{dyngraph2vecAERNN~\cite{goyal2020dyngraph2vec}}: This method is similar to dyngraph2vecAE, except that the feed-forward neural network in the encoder is replaced by LSTM layers. This method also incorporates historical information while predicting graph embeddings.
    \item \textit{EvolveGCN~\cite{pareja2020evolvegcn}}: A graph convolution network (GCN) forms the core of this model. A recurrent neural network (RNN) is used to evolve the GCN parameters along time.
    \item \textit{ROLAND~\cite{you2022roland}}: A graph neural network is combined with GRU to learn the temporal dynamics. The GNN has attention layers to capture the spatial-level node importance. For this case, we ran the code from the official GitHub repository \footnote{\url{https://github.com/snap-stanford/roland}} 5 times and the mean and standard deviation of the MAP and MRR are reported. We note the original code did not include MAP computation so we added to the code for comparison with our benchmarks. We also note that the results reported in \cite{you2022roland} are slightly different from our results because the average in \cite{you2022roland} was taken over only 3 times. The reason we ran it 5 times is because we observed a large variability in the results of the individual runs.
    \item \textit{DynG2G~\cite{xu2022dyng2g}}: This method uses the G2G encoder at its core to obtain the embeddings. The weights from the previous timestamp are used to initialize the weights of the next timestamp to make the training faster. Unlike the previous methods, DynG2G includes information about uncertainty of the embeddings.
\end{itemize}
We see that the TransformerG2G model outperforms other methods for all datasets except SBM and Slashdot dataset. The MAP and MRR values for the SBM dataset of TransformerG2G is very close to that of DynG2G and better than other methods. We did not see an improvement in MAP here due to the Markovian nature of the dataset. In the case of Slashdot, TransformerG2G did not perform well because of the low number of timestamps and the high novelty of the dataset. Note that in the multi-step method, the importance given to different historical timestamps is a hyperparameter chosen by the user,  whereas in TransformerG2G, the attention mechanism at its core adaptively assigns the importance to the different timestamps. For all the baseline methods, except ROLAND, we have used the MAP and MRR values as reported in \cite{xu2022dyng2g} and \cite{pareja2020evolvegcn}. However, for the Reality Mining and Slashdot datasets, the MAP and MRR values for all the baselines were not available in the referenced resources or in the articles that introduced these baseline models,  hence we did not compare against all the other baseline methods.

\begin{table}[!h]
\caption{Comparison results of temporal link prediction task for six different benchmark datasets.}
\label{tab:method_comparisons_LP}
\centering
\fontsize{6.6}{8}\selectfont
\begin{tabular}{llll}
\hline
Benchmark              & Method                     & MAP             & MRR    \\
\hline \hline

\multirow{5}{*}{SBM} & DynGEM                     & 0.1680           & 0.0139 \\
                     & dyngraph2vecAE             & 0.0983          & 0.0079 \\
                     & dyngraph2vecAERNN          & 0.1593          & 0.0120  \\
                     & EvolveGCN                  & 0.1989          & 0.0138 \\
                     & DynG2G$^*$            &  \textbf{0.6433 $\pm$ 0.0187}   & \textbf{0.0409 $\pm$ 0.0014}\\
                     & TransformerG2G ($l = 1$)   & $0.6204 \pm 0.0386$        & 0.0369 $\pm$ 0.0055\\
\hline
\multirow{2}{*}{Reality Mining}   

                                & EvolveGCN                  & 0.0090          & 0.0416 \\
                                & DynG2G$^*$  &  0.1126 ± 0.0322  & \textbf{0.1973 ± 0.0341}                   \\
                                & TransformerG2G ($l = 4$)   & \textbf{0.2252 ± 0.0137}         & 0.1294 ± 0.0129\\
\hline
\multirow{5}{*}{UCI}         & DynGEM                     & 0.0209          & 0.1055                     \\
                             & dyngraph2vecAE             & 0.0044          & 0.0540                     \\
                             & dyngraph2vecAERNN          & 0.0205          & 0.0713                     \\
                             & EvolveGCN                  & 0.0270          & 0.1379                     \\
                             & ROLAND                     & $0.0080 \pm 0.0081$  & $0.11844\pm 0.0849$ \\
                             & DynG2G$^*$                     & $0.0348\pm 0.0121$          & $0.3247\pm 0.0312$                    \\
                             & TransformerG2G ($l = 5$)   & \textbf{0.0495 ± 0.0107}         & \textbf{0.3447 ± 0.0196}\\
\hline
\multirow{1}{*}{Slashdot}  & DynG2G$^*$                  & \textbf{0.1012 ± 0.0053}         & \textbf{0.3885 ± 0.0030}                     \\
& TransformerG2G ($l = 1$)   & 0.0498 ± 0.0059         & 0.2568 ± 0.0126\\
\hline
\multirow{5}{*}{Bitcoin-OTC} & DynGEM                     & 0.0134          & 0.0921                     \\
                             & dyngraph2vecAE             & 0.0090          & 0.0916 \\
                             & dyngraph2vecAERNN & 0.0220 & 0.1268                     \\
                              & EvolveGCN                  & 0.0028          & 0.0968                     \\
                              & ROLAND        & $0.0014 \pm 0.0010$  & $0.1863 \pm 0.0675$ \\
                             & DynG2G$^*$                   & 0.0083 ± 0.0027       &0.3529 ± 0.0100                         \\
                             & TransformerG2G ($l = 4$)   & \textbf{0.0303 ± 0.0042}         &  \textbf{0.4037 ± 0.0188}\\
\hline
\multirow{5}{*}{AS}  & DynGEM                     & 0.0529          & 0.1028 \\
                     & dyngraph2vecAE             & 0.0331          & 0.1028 \\
                     & dyngraph2vecAERNN          & 0.0711          & 0.0493 \\
                     & EvolveGCN                  & 0.1534          & \textbf{0.3632} \\
                     & DynG2G$^*$            & 0.3139 ± 0.0129   & 0.3004 ± 0.0040  \\
                     & TransformerG2G ($l = 1$)   & \textbf{0.3976 ± 0.0594}         & 0.3599 ± 0.0276\\
\hline \hline
\end{tabular}
\end{table}

% \begin{figure}[H]
%     \centering
%     \includegraphics[width = \textwidth]{figure/map_plot.pdf}
%     \caption{MAP values for temporal link prediction based on node embeddings learned with our proposed TransformerG2G model across six dynamic graph datasets, where higher MAP values indicate better link prediction performance. The datasets corresponding to a)-f) are SBM, Reality Mining, UCI, Bitcoin, Slashdot, and Autonomous Systems, respectively. The blue bars display the MAP values achieved by DynG2G, while the yellow bars present the MAP values obtained by our TransformerG2G model with different lookbacks ($l = \{1,2,3,4,5\}$).}
%     \label{fig: MAP_plot}
% \end{figure}

\subsection{Time-dependent attention matrix}

\begin{figure}[!ht]
    \centering
    \includegraphics[width = 0.93\textwidth]{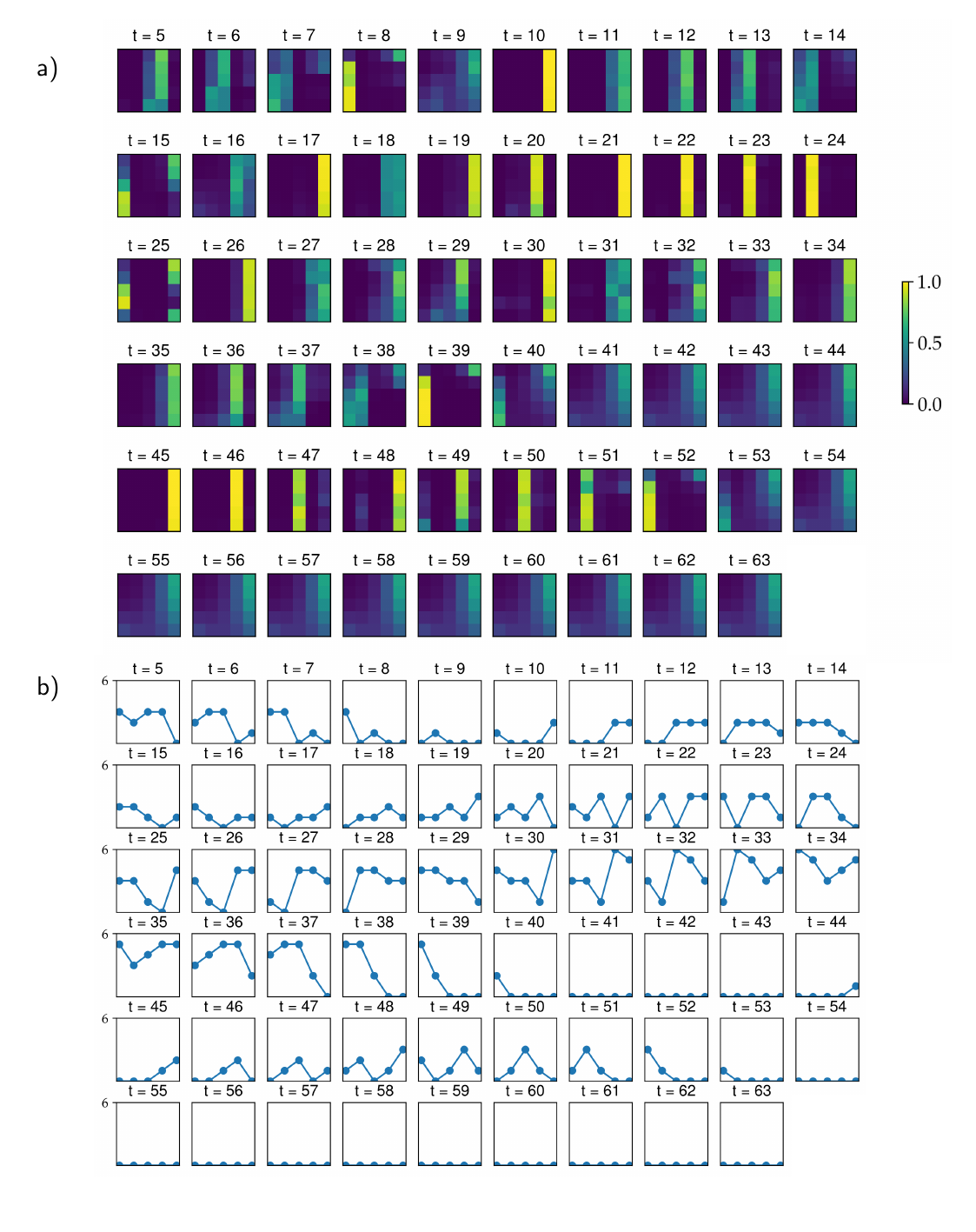}
    \caption{The attention matrix of TransformerG2G with $l = 4$ for a randomly selected node (ID No. 6) in Reality Mining is visualised at different time stamps in a). The attention matrix shows the importance given to the current state and previous context while generating the graph embeddings. b) shows the node degree of the current state and previous context. Interestingly, we see that the learned model assigns more significance to time stamps where the node has a higher degree, indicating that the model recognized the importance of nodes with greater connectivity.}
    \label{fig: Attention}
\end{figure}

We visualize the attention matrices (i.e., $softmax(QK^T/\sqrt{d})$ 
% ({\color{red} Mengjia: add this formula and definitions of Q, K, d parameters in section 3.3.}) 
of the trained TransformerG2G model in to gain insights about which timestamps are prioritized by the transformer while predicting the embeddings. Figure \ref{fig: Attention} a) shows the attention matrix for a randomly selected node (ID No.6) from the Reality Mining dataset at different timestamps from the transformer model with $l = 4$. The \textit{[4-$i$,4-$j$]}-th entry in the attention matrix corresponding to timestamp $t$ denotes the temporal feature importance given to the representation of the selected node at timestamp $t-j$ by the representation of the same node at timestamp $t-i$. To further comprehend the patterns in the attention matrices, we show the node degree of the current timestamp and its context in Fig.\ref{fig: Attention}. We see that the model assigns more importance to timestamps where the node has higher connectivity (higher node degree). From timestamps 54 to 63, the selected node is absent from the graph since it has node degree 0 as evident from Fig \ref{fig: Attention} b). We can see the reflection of this in the attention matrices from timestamps 54 to 63. The attention matrices do not assign significance to a particular historical timestamp and is confused which historical timestamp it should prioritize.
% Attention matrix plot

\section{Summary} \label{conclusions}
We presented the multi-step DynG2G and TransformerG2G deep learning models to learn embeddings in dynamic graphs. Our models leverage information about the temporal dynamics in the graph predicting embeddings, and also quantify the uncertainty associated with the node embeddings. In the multi-step method, we use a weighted scheme to initialize the weights of deep learning. The TransformerG2G model, based on the transformer architecture, uses the attention mechanism to adaptively give importance to the previous timestamps. We conducted experiments with varying lookbacks $l$, and found that the TransformerG2G model achieves better performance on link prediction task on six benchmark datasets. We also showed that the temporal evolution of the attention matrices in the TransformerG2G model follows the evolution of the node degree, and thereby pays attention to timestamps where the node has a higher connectivity.

One of the limitations of this work is that we use the same embedding dimension during all timestamps. Using an adaptive embedding dimension that is optimal for each timestamp could improve the quality of the learned embeddings. This could further lead to better performance in the downstream link prediction tasks. In this work, we used the vanilla positional embedding and scaled dot-product attention within the transformer. It might be useful to explore other attention mechanisms and positional encoding methods. Future works could also look at the extension of the attention-based methods for learning embeddings in hyperbolic space.

\section*{Acknowledgements}
We would like to acknowledge support by the DOE SEA-CROGS project (DE-SC0023191) and the ONR Vannevar Bush Faculty Fellowship (N00014-22-1-2795).

\newpage
%-------------- References -------------------%
%%
%% Following citation commands can be used in the body text:
%% Usage of \cite is as follows:
%%   \cite{key}         ==>>  [#]
%%   \cite[chap. 2]{key} ==>> [#, chap. 2]
%%

%\bibliographystyle{elsarticle-num}
%\bibliography{ref}

\end{document}